\documentclass[lettersize,journal]{IEEEtran}
\usepackage{amsmath,amsfonts,amssymb,mathtools,bm}
\usepackage{fontspec}
\usepackage{array,booktabs,tabularx,graphicx,stfloats,url}
\usepackage[nocompress]{cite}
\usepackage[hidelinks,bookmarksnumbered=true]{hyperref}
\newcolumntype{Y}{>{\raggedright\arraybackslash}X}
\newcolumntype{P}[1]{>{\raggedright\arraybackslash}p{#1}}

\hypersetup{pdftitle={Training-Free Interventions for Hallucination in Large Language Models},pdfauthor={Can Huang}}
\begin{document}
\bstctlcite{IEEEcontrol}
\title{Training-Free Interventions for Hallucination in Large Language Models}
\author{Can Huang%
\thanks{Can Huang is with the School of Artificial Intelligence, Beijing University of Posts and Telecommunications, Beijing, China (e-mail: \mbox{\href{mailto:huangcan@bupt.edu.cn}{huangcan@bupt.edu.cn}}).}}
\markboth{Training-Free Hallucination Intervention}{Can Huang}
\maketitle
\begin{abstract}
A frozen language model can give a wrong answer because a fact is unavailable, a reliable passage is ignored, or an unsupported claim survives generation. We review 116 works on inference-time hallucination intervention and its foundations. We trace the development from retrieval and controllable generation to methods that act on input evidence, token scores, internal representations, completed claims, or the decision to answer. For each family we compare its source of evidence, model access, preparation cost, and the task on which a correction can be established. For activation steering, we distinguish estimating a subspace from applying a state-dependent edit and separate interventions on the prompt from those on the generated prefix. Published comparisons show why faithfulness to a source, factual correctness, informativeness, and answer coverage must be measured separately. Applications in time-sensitive question answering, document synthesis, and visual grounding identify when source authority, evidence availability, or capability preservation limits a reported gain.
\end{abstract}
\begin{IEEEkeywords}
Hallucination, evidence grounding, inference-time intervention, activation steering, factuality, selective answering.
\end{IEEEkeywords}
\section{Introduction}\label{sec:intro}

Suppose a passage gives the previous holder of an office and a question asks who holds it today. Copying the name is faithful to the passage. Giving a newer name may be factually correct but unsupported by that passage. If the question instead asks who held office according to the document, the appropriate answer changes again. This distinction, central to earlier reviews of grounded generation and LLM factuality~\cite{ji2023survey,huang2025survey}, determines what an intervention should reward. A source-conditioned summary, an open-domain question, and a request to resolve conflicting reports require different decisions even when the model produces the same words.

\begin{figure*}[!t]
\centering
\includegraphics[width=\textwidth]{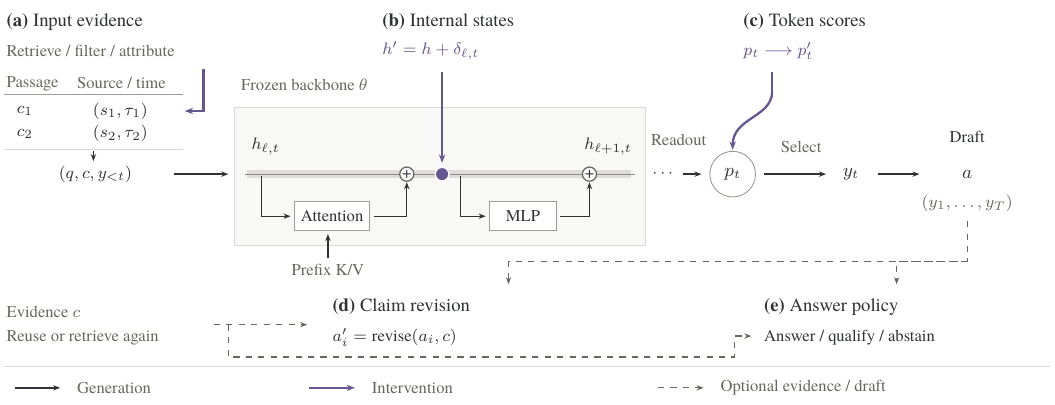}
\caption{Five families located by their intervention objects: evidence $c$, hidden states $h$, token probabilities $p_t$, claims $a_i$, and the decision to answer. The current token is shown in one decoder layer, with other layers and normalization omitted; attention also reads prefix keys and values. Solid arrows follow generation, purple arrows mark edits, and dashed arrows show optional use of evidence or a draft. The families may be used separately or combined.}\label{fig:sites}
\end{figure*}

A frozen language model can acquire a new observation through search, use an existing observation differently, or revise an unsupported assertion after it has been made. These interventions operate on different information. A passage can settle a question about an unfamiliar event. Reading the same passage a second way can reveal that the decoder ignored its date. Repeatedly sampling the same model can expose disagreement about the event, although agreement alone supplies no independent witness. Figure~\ref{fig:sites} places these choices along the path to an answer. Two questions determine the choice: what evidence would settle the task, and at what point it can still affect the answer~\cite{jiang2023flare,shi2024cad,manakul2023selfcheck,gao2023rarr}.

Reviews of natural language generation and large language models distinguish factual errors from failures of faithfulness and trace their causes through data, training, and inference~\cite{ji2023survey,huang2025survey,zhang2025siren}. Surveys of knowledge conflict and visual hallucination refine what counts as disagreement between sources~\cite{xu2024conflicts,sahoo2024survey}. Our scope is narrower in timing and broader in the kinds of decisions compared. We examine what can be changed for a particular request after the backbone has been trained, from the evidence supplied to the decision to withhold an answer. General steering results enter the review where they clarify a mechanism or a cost. A change in sentiment or style alone provides no evidence of factual improvement.

The review synthesizes 116 works. Section~\ref{sec:landscape} traces changes in the research questions and introduces a common classification. Sections~\ref{sec:evidence}--\ref{sec:verification} compare the evidence and access required by each family. Sections~\ref{sec:evaluation} and~\ref{sec:applications} examine how to interpret experiments and how the criterion for success changes across tasks. Section~\ref{sec:challenges} examines when to override model memory or correct a deviation across layers. The appendix describes corpus selection.

\section{Development and Taxonomy}\label{sec:landscape}

\begin{figure*}[!t]
\centering
\includegraphics[width=\textwidth]{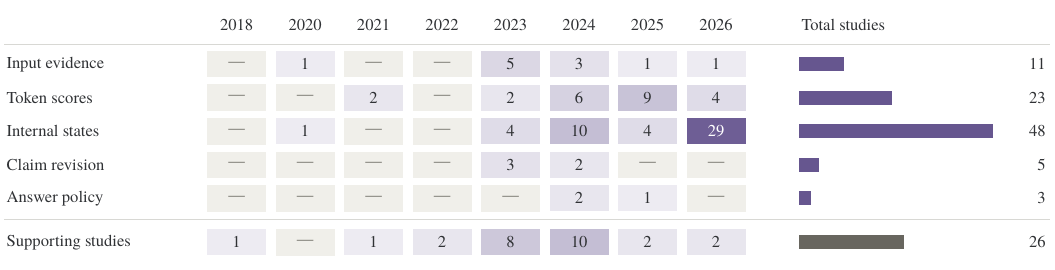}
\caption{Selected literature by year and primary family. Each of 116 works is counted once. Cell numerals give exact counts, darker cells indicate larger counts, and dashes denote zero; bars give family totals. Supporting studies include benchmarks, surveys, and foundations. Year follows the cited publication or preprint record.}\label{fig:corpus}
\end{figure*}

\subsection{From adding evidence to controlling its use}

Two early lines of work addressed different obstacles. Retrieval-augmented generation made an external collection available as a source of facts, with learned retrieval and generation components~\cite{lewis2020rag}. Controllable generation showed that a frozen model could change its continuation when guided by activation gradients, auxiliary token distributions, or a predictor of a future attribute~\cite{dathathri2020pplm,liu2021dexperts,yang2021fudge}. A third result connected them: when a passage contradicted model memory, the answer could depend on the entity as well as on the supplied text~\cite{longpre2021knowledge}. Access to an external memory and the ability to steer a decoder were distinct technical advances. Neither, by itself, defined which source should prevail.

By 2023--2024 the point of intervention had become a design choice within the same task. A system could retrieve during reasoning, contrast token probabilities with and without context, alter selected attention heads, or check a finished claim against retrieved text~\cite{trivedi2023ircot,shi2024cad,li2023iti,gao2023rarr}. These alternatives exposed different bottlenecks. A missing bridge fact rewards a better query. A bridge fact already in the prompt rewards better use of evidence. If the only error is an unsupported modifier in an otherwise correct answer, a local revision preserves more of the useful work. This distinction explains why advances in retrieval did not settle the decoding or revision problem.

Recent work makes the reliability and cost of an edit explicit. Source labels help separate retrieved from generated evidence, perturbations to retrieval heads probe prefix dependence, and state-dependent operators can restrict a change to selected components~\cite{wang2025astute,gema2025decore,luo2026skop,ghasemi2026orbit}. Studies of shared decoding directions and capability trade-offs ask what the same intervention removes from other decisions~\cite{decodeshare2026,sharma2026lowrank,raj2026frontier}. Figure~\ref{fig:corpus} locates these strands in the selected literature. Of 48 works assigned to internal computation, 29 fall in 2026 by the recorded year. This concentration reflects the review’s emphasis on activation steering. It motivates a closer examination of geometry and timing without implying that most hallucination research concerns internal states.

\subsection{Intervention sites and preparation costs}

Figure~\ref{fig:sites} classifies methods by the object on which they implement the correction. A system that retrieves a passage and then verifies its draft spans two families. A decoder that masks heads solely to construct a reference distribution is grouped with decoding because the resulting contrast selects the token. By contrast, a head edit deployed directly in the ordinary forward pass belongs to representation control. Following the causal path of a correction makes these assignments interpretable even when two methods use the same module~\cite{gema2025decore,li2023iti,gao2023rarr}.

\begin{figure}[!t]
\centering
\includegraphics[width=\columnwidth]{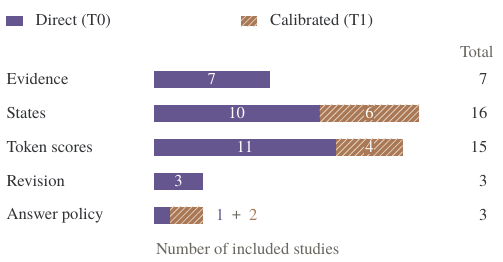}
\caption{Preparation in the 44 core studies, counted once by primary family. Bar segments show direct use of existing components (T0, 32 papers) and calibration statistics or validation choices (hatched T1, 12). Labels give segment counts; the rightmost column gives family totals.}\label{fig:preparation}
\end{figure}

Preparation is an independent axis. T0 deploys existing components directly. T1 estimates statistics, a threshold, or a basis from calibration examples without fitting a method-specific predictor. T2 fits an auxiliary probe, controller, or retriever. T3 updates the target backbone. T0 and T1 form the core here. For example, subtracting example activations, averaging labeled contrasts, and using a fitted probe to select intervention heads can all lead to the addition of a vector at deployment~\cite{turner2023actadd,rimsky2024caa,li2023iti}. Learned low-rank interventions and corrections amortized into model components have further preparation requirements~\cite{wu2024reft,yang2026casal}. These distinctions matter for replication: access to model weights, labeled calibration examples, an external search service, and repeated forward passes are separate resources.
Figure~\ref{fig:preparation} restricts the count to core studies. Six of 16 studies of internal states and four of 15 token-scoring studies use calibration. For all seven input-evidence studies the core variant uses existing components directly. A frozen backbone still leaves an important replication question: whether reproducing the intervention requires fresh calibration data. ITI and SPACE fit auxiliary probes and are counted as T2 comparators~\cite{li2023iti,wang2025space}.

Table~\ref{tab:methods} compares representative implementations at their deployment interfaces. Preparation codes refer to the stated variant: selecting settings from held-out labeled examples adds calibration cost even when the deployed operator has no learned parameters. CCS is included as an enabling steering method, not as direct evidence of hallucination reduction.

CDA is counted under token scores and cross-referenced under answer policy. Conformal factuality is assigned to answer policy because it calibrates how much of a draft to assert; its claim-level editing also connects it to revision. These secondary roles preserve the relationship between mechanisms without counting a paper twice.

DoLa is coded T1 here because the reported protocol selects the candidate layer bucket on validation data; token-wise layer selection then operates within that bucket~\cite{chuang2024dola}.

\begin{table*}[!t]
\centering
\caption{Representative methods: required access, preparation, and deployment cost}\label{tab:methods}
\footnotesize\renewcommand{\arraystretch}{1.13}\setlength{\tabcolsep}{4pt}
\begin{tabularx}{\textwidth}{@{}P{.16\textwidth}P{.17\textwidth}P{.045\textwidth}P{.255\textwidth}Y@{}}
\toprule
\textbf{Method} & \textbf{Edit / model access} & \textbf{Prep.} & \textbf{Extra work or evidence at inference} & \textbf{Main limitation} \\
\midrule
FreshPrompt~\cite{vu2024freshllms} & Input / text API & T0 & Search results and a longer evidence prompt & Retrieved statements can be stale or misleading. \\
CAD~\cite{shi2024cad} & Token scores / logits & T0 & Predictions with and without context & More context reliance need not mean more truth. \\
DoLa~\cite{chuang2024dola} & Token scores / layer readouts & T1 & Intermediate readouts and layer comparison & Layer disagreement is not a factuality label. \\
CDA (basic)~\cite{kim2025cda} & Token scores / logits & T0 & Memory, context, abstention, and null-prompt branches & Confidence depends on prompts; extra decoding work. \\
SEA~\cite{qiu2024sea} & Internal states / activation hooks & T1 & Apply estimated projection operators & Depends on calibration data and edit locations. \\
CCS$^{\dagger}$~\cite{xenos2026ccs} & Internal states / activation hooks & T1 & Add a subspace-filtered fixed vector & Projection alone does not create a dynamic edit. \\
RARR~\cite{gao2023rarr} & Draft claims / text API & T0 & Evidence search, agreement checks, and revision & A supported edit can change the original intent. \\
CoVe~\cite{dhuliawala2024cove} & Draft claims / text API & T0 & Verification questions and a revised response & Self-verification still shares model knowledge. \\
Conformal abstention~\cite{abbasi2024conformal} & Answer policy / text API & T1 & Sample responses and assess their similarity & Risk control depends on the calibration setting. \\
Conformal factuality~\cite{mohri2024conformal} & Retained claims / text API & T1 & Score subclaims and reduce output specificity & Marginal validity does not ensure informativeness. \\
\bottomrule
\end{tabularx}
\par\smallskip
\parbox{\textwidth}{\footnotesize T0: direct use of existing components; T1: sample-based calibration or subspace estimation. Costs describe operations, not comparable latency measurements. $^{\dagger}$General behavioral steering included as enabling evidence.}
\end{table*}

\section{Evidence Acquisition and Source Arbitration}\label{sec:evidence}

Retrieval changes what the generator can know about a request. Its effect passes through three decisions: which evidence to seek, which source to trust, and which of the available passages the generator actually uses. Failure at any of these points can produce the same wrong answer. They can be distinguished by supplying a verified passage directly, replacing it with a plausible false passage, and moving it within an otherwise identical prompt~\cite{longpre2021knowledge,liu2024lostmiddle,wang2025astute}.

\subsection{Fixed, iterative, and anticipatory retrieval}

A fixed query works when the question names the entity and relation needed to retrieve the answer. FreshQA illustrates why time and premise are also part of that query. It separates changing facts and false-premise questions, and its prompting method organizes search results for a frozen generator~\cite{vu2024freshllms}. For a compositional question, the missing item may be the entity linking two facts. Self-Ask exposes that item as a subquestion. ReAct can call a tool after an intermediate reasoning step. Interleaving retrieval with reasoning uses the evolving chain to retrieve the next supporting passage~\cite{press2023selfask,yao2023react,trivedi2023ircot}. Each lets a later query depend on newly obtained information. A mistaken bridge entity can also send the remaining searches toward a coherent answer about the wrong subject.

The unit used to trigger a search matters. An anticipated sentence exposes particular low-confidence tokens and can be regenerated after retrieval. A complete draft carries more relationships that can aid a new search, at the cost of carrying more unsupported assumptions~\cite{jiang2023flare,shao2023iterretgen}. A learned retrieval evaluator or a model trained to emit reflection tokens can decide when to change strategy~\cite{yan2024crag,asai2024selfrag}. This moves part of the decision into training. Comparing adaptive policies therefore requires the same document collection, separate evidence-recall and answer scores, and a count of search and model calls. Otherwise a gain from finding the missing passage is indistinguishable from a gain in using the very same passage.

\subsection{Conflicts between memory and retrieved accounts}

Synthetic entity replacements allow the model's usual answer and a document's answer to be varied independently~\cite{longpre2021knowledge}. A second line constructs a coherent counter-account rather than swapping a name. Its authors find models highly receptive when this is the only external account, and attribute part of the difference from entity-replacement results to the persuasiveness of the evidence~\cite{xie2024conflict}. A reported preference for memory or context thus depends on how the conflict was written. Probes can reveal which source a residual state favors~\cite{residual2024conflict}, while changing an estimated knowledge-selection direction can alter which answer is expressed~\cite{lou2026ksd}. Neither observation supplies the ground-truth authority. In a counterfactual reading task the supplied passage defines the answer. In a factual update task a dated passage requires a freshness check. In multi-document QA, two retrieved passages may disagree even if the model has no strong prior~\cite{xu2024conflicts}.

ASTUTE RAG keeps retrieved passages distinguishable from model-generated accounts, groups compatible claims, and resolves disagreement with provenance still visible~\cite{wang2025astute}. This addresses a weakness of treating repetition as corroboration. Several retrieved copies may derive from one article, and several generated accounts may repeat one memorized error. Local attribution offers a second design choice. Selecting identifiable spans and grouping them before drafting a sentence preserves a trace from the sentence to its sources~\cite{slobodkin2024attribute}. End-to-end citation evaluation then asks both whether the cited material covers the statement and whether each citation contributes support~\cite{gao2023alce}. Provenance permits these checks even when the reliability of the underlying source remains unsettled.

\begin{figure}[!t]
\centering
\includegraphics[width=\columnwidth]{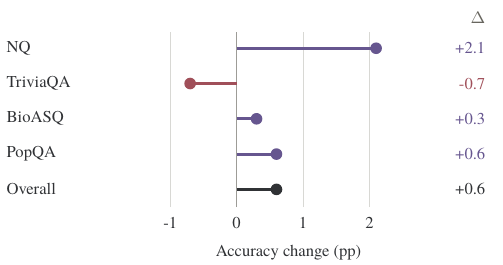}
\caption{Accuracy with source labels minus accuracy without them, in percentage points. The zero line marks no effect. Overall is the aggregate reported in Table 3 of ASTUTE RAG. Differences use the rounded accuracies in that table~\cite{wang2025astute}.}\label{fig:source}
\end{figure}

The source-label ablation in Fig.~\ref{fig:source} shows the size and variability of this effect. Keeping provenance improves reported accuracy by 2.1 points on Natural Questions and lowers it by 0.7 on TriviaQA, with a 0.6-point overall gain in the same experiment~\cite{wang2025astute}. The within-system comparison isolates the presence of labels more closely than comparing two complete retrieval systems. Its variation across datasets leaves open when the model can use those labels to identify an authoritative passage.

\subsection{Position and attention as limits on evidence use}

A longer prompt can contain the answer and still conceal it functionally. Moving a relevant passage within the context window changes performance on long-context tasks~\cite{liu2024lostmiddle}. Reweighting marked spans at selected heads tests whether the model can be directed to that position~\cite{zhang2024pasta}. A classifier that masks tokens judged unreliable makes a stronger assumption about their truth~\cite{yu2024tacs}. The three manipulations isolate different causes. Moving a passage tests position sensitivity, emphasizing it tests accessibility, and removing it trusts a source-selection rule. A strong grounded evaluation holds the passage fixed across these interventions and includes a comparison in which the emphasized passage is wrong.

\section{Token Selection During Decoding}\label{sec:decoding}

Once the evidence is fixed, two models or two passes through one model can still favor different continuations. A contrastive decoder treats this disagreement as information about a candidate token. Its behavior is determined by what differs between the passes, how their scores are combined, and whether the intervention can revisit an earlier choice.

\subsection{Reference distributions and admissible candidates}

One way to use evidence already present is to ask which candidates lose probability when some source of information is removed. For a vocabulary item $w$, let $p_+$ be the preferred prediction and $p_-$ the reference. Many contrastive decoders can be compared through the score in (\ref{eq:score}), applied over their allowed candidates $\mathcal V_t$. A word equally probable in both branches receives no boost from their log-ratio. A word favored only by the preferred branch rises relative to its competitors. The weights and admissible vocabulary are part of the method. Without a plausibility constraint, a very small reference probability can promote a word the preferred model itself regards as implausible~\cite{li2023contrastive}.

\begin{equation}
s_t(w)=a_t\log p_+(w)-b_t\log p_-(w),\quad w\in\mathcal V_t.\label{eq:score}
\end{equation}

\begin{figure*}[!b]
\centering
\includegraphics[width=\textwidth]{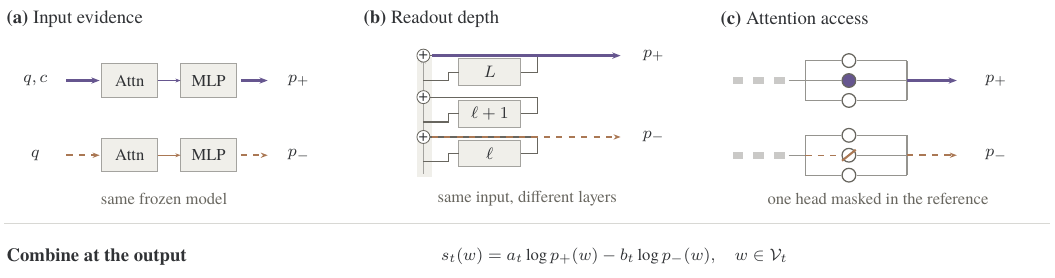}
\caption{Three reference predictions at the same generation step. Left: the same frozen model is conditioned on $q,c$ or $q$. Center: earlier and final readouts share a residual stream. Right: masking selected heads changes access to the prefix. The paired scores are combined at the output; weights and the admissible vocabulary are method-specific~\cite{shi2024cad,chuang2024dola,gema2025decore}.}\label{fig:branches}
\end{figure*}

Three choices of reference answer different questions (Fig.~\ref{fig:branches}). Context-aware decoding (CAD) removes the context to estimate the contribution of the passage relative to the model prior~\cite{shi2024cad}. Masking some input tokens or disrupting lexical or syntactic information measures sensitivity to a particular corruption~\cite{huang2025delta,chen2026pmi}. Comparing earlier and later readouts keeps the input fixed. DoLa contrasts the final layer with a selected earlier layer; SLED instead adjusts logits using their evolution across layers~\cite{chuang2024dola,zhang2024sled}. Growth across depth can reflect successful integration of a fact, an instruction, or an increasingly confident mistake. An experiment that changes both the passage and the readout cannot identify which of these signals drove the result.

A third family builds a deliberately weakened prediction. DeCoRe masks retrieval heads to reduce access to earlier tokens. Selecting important heads by gradient and dispersing their attention changes a different set of dependencies~\cite{gema2025decore,jiang2025hicd}. An induced hallucinating branch can arise from prompting or from a separately fitted negative model. A self-induced reference can instead use particular layers, while an adversarial low-rank negative branch requires fitting additional parameters~\cite{zhang2025icd,jing2026shicd,sheng2025rcd}. Earlier expert/anti-expert mixtures and future-attribute predictors explain how such scores can favor a behavior learned in advance~\cite{liu2021dexperts,yang2021fudge}. Their ability to discriminate that behavior depends on what the negative branch lost. It may have lost ordinary language competence together with the unwanted bias.

\subsection{Adaptive strength, abstention, and recovery}

Uncertainty can trigger a new score, a new search, or a return to an earlier prefix. Entropy-weighted contrast changes the strength of the present step~\cite{gema2025decore}. A learned policy can turn layer contrast on selectively, with an offline reinforcement-learning cost~\cite{zhang2025actlcd}. Backtracking approaches use confidence or distance to the source to reconsider earlier token choices~\cite{liu2023coba}. The unit of repair matters once an entity has been introduced. Reweighting its continuation may yield a fluent sentence about the wrong person, whereas undoing the prefix or retrieving a new passage can change the entity itself. Such comparisons need a matched token or latency budget, since a method allowed to revisit an error receives opportunities denied to a single forward pass.

Abstention can also be encoded in the token score. Contrastive Decoding with Abstention (CDA) combines logits from memory-only, context-conditioned, and explicitly abstaining prompts~\cite{kim2025cda}:
\begin{equation}
z_t=w_t^{p}z_t^{p}+w_t^{c}z_t^{c}+(1-w_t^{p}-w_t^{c})z_t^{a}.\label{eq:cda}
\end{equation}
The weights use confidence signals relative to content-free prompts; they do not certify that a source is true. Basic CDA constructs these signals at inference time (T0), whereas its momentum variant adds a smoothing hyperparameter. Multiple prompt branches increase decoding work. This connects output control to the answer policy in Section~\ref{sec:verification}: when neither knowledge source is usable, the available action need not be a forced answer.

The distinction between grounding and truth is especially sharp at the token level. If a passage contains a wrong date, the token whose probability drops most when the passage is removed may be exactly that date. Context contrast can then increase faithfulness and increase factual error on the same example. Testing the branch on correct passages, misleading passages, and passages that omit the answer separates sensitivity to evidence from reliability under conflict~\cite{shi2024cad,longpre2021knowledge,xie2024conflict}.

\section{Control of Internal Representations}\label{sec:representations}

Activation steering intervenes before the output distribution is formed. A direction estimated at one layer can alter later attention, the residual stream, and the eventual word choice. To compare interventions one must specify how the relevant feature was identified, which transformation acts on a live state, and at which layers and token positions the transformation runs. The same estimated direction can support several different interventions under these choices.

\begin{figure*}[!t]
\centering
\includegraphics[width=\textwidth]{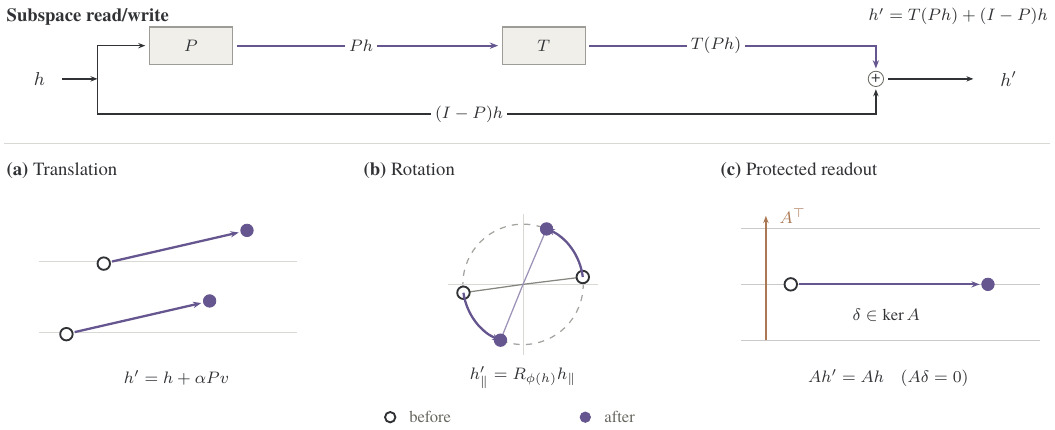}
\caption{Subspace read/write and three edits. Above, only $Ph$ is transformed and then recombined with the unchanged complement $(I-P)h$. Below, two-dimensional examples show parallel fixed displacements, state-dependent norm-preserving rotations, and edits satisfying $A\delta=0$ that preserve a scalar linear readout $Ah$. These local constraints do not guarantee unchanged downstream decisions.}\label{fig:operators}
\end{figure*}

\subsection{Feature identification and intervention sites}

A representation that predicts a truth label is a candidate location for intervention. Trained probes can discriminate statements, geometric analyses can identify directions that generalize across some controlled datasets, and a contrast-consistency objective can discover features without truth labels~\cite{azaria2023internal,marks2024geometry,burns2023ccs}. These findings concern readout. An intervention makes a stronger claim about what happens when the feature is changed. Useful controls include matched statements with reversed labels, paraphrases that preserve the fact, and interventions of the same magnitude in random directions. They address, respectively, confounding by sentence form, sensitivity to phrasing, and generic disruption. In fact, the geometry-of-truth study intervenes in both true-to-false and false-to-true directions and measures the normalized change in the model’s labeling decision~\cite{marks2024geometry}. That is more specific evidence about causal use than classification accuracy alone. A probe that predicts whether a context will win a conflict~\cite{residual2024conflict} leaves open whether changing that feature will select the reliable source.

Additive steering estimates a change from examples and applies it to new states. A contrast between prompts, an average over paired responses, and a vector distilled from demonstrations differ in how much nuisance variation can be averaged away~\cite{turner2023actadd,rimsky2024caa,liu2024icv}. Representation engineering generalizes this distinction between reading a concept and controlling behavior~\cite{zou2023repe}. Targeted head interventions use separate evidence to choose where to intervene, sometimes fitting a probe even when the deployed displacement is a mean difference. Nonlinear probe and multi-token extensions change the information available for selection and the locations being treated~\cite{li2023iti,hoscilowicz2024nliti}. Thus a gain attributed to a direction may actually depend on the examples, the selected heads, or the positions at which it is applied. A useful ablation exchanges one of these choices at a time.

A residual edit changes information available to subsequent modules. An attention edit can instead change which tokens contribute to the current update. Retrieval-head ablations locate components needed to recover distant tokens~\cite{wu2025retrievalhead}. A vocabulary-derived semantic basis can localize heads and constrain directions for predefined attributes~\cite{tallini2026locus}. In visual generation, context-dependent tilting reweights image regions according to the current token~\cite{lei2026cai}. When moving between these settings, the semantic target must be re-established. A dictionary that captures topical or affective words need not distinguish a valid recent update from a well-written false correction.

\subsection{Directions, subspaces, and protected quantities}

A subspace can enter the analysis at three different stages. It can remove nuisance components from an estimated direction, define the part of a live state to transform, or constrain which quantities may change. Let $B$ have orthonormal columns and $P=BB^\top$. In (\ref{eq:space}), $\delta_1$ filters a fixed displacement $v$, while $\delta_2$ permits the coefficient vector $a(h)$ to depend on the state. Even with a large basis, changing only a scalar gain on $Pv$ yields one direction of motion. A genuinely multidimensional edit requires the coefficients or the operator acting on the current state to vary in how they weight basis directions. These distinctions explain why papers reporting similar ranks can test different hypotheses.

\begin{equation}
\delta_1=\alpha Pv,\qquad \delta_2(h)=B\,a(h),\quad a(h)\in\mathbb R^{r}.\label{eq:space}
\end{equation}

The choice of data gives a basis its meaning. Contrastive Covariance Subspaces (CCS) contrasts the scatter of a target concept with that of other concepts to suppress topic or style shared across the collection before projecting a fixed steering vector~\cite{xenos2026ccs}. Spectral Editing of Activations (SEA) instead decomposes cross-covariance between neutral and positive or negative activations to construct operators on live representations~\cite{qiu2024sea}. Soft projection matrices can attenuate different components continuously and can be composed across concepts~\cite{postmus2024conceptor,triantafyllopoulos2026conceptor}. A learned semantic/truth latent space or learned probes for joint truthfulness and faithfulness move the separation into an auxiliary model~\cite{zhang2024truthx,wang2025space}. For all of these, a high-variance direction inherits the composition of the calibration examples. Transfer to a new topic, model, or stage must be tested separately from fit to the calibration data.

Rotation provides another use of a subspace. A fixed plane supports an angular edit. ORBIT combines attribute planes with input-dependent gates, allowing the displacement to depend jointly on the token and the requested attributes~\cite{vu2025angular,ghasemi2026orbit}. Pure rotation preserves the norm within its plane, although an unrelated token can cross a decision boundary. Steering via Key-Orthogonal Projections (SKOP) instead estimates query directions along which key differences are especially likely to change attention and restricts the edit in those directions~\cite{luo2026skop}. Figure~\ref{fig:operators} contrasts three geometries. Equality of a norm or a selected linear readout can be checked locally. Preservation of ordinary question answering must be measured after the remaining layers have run.

\subsection{Prompt processing, cached decoding, and depth}

\begin{figure}[!t]
\centering
\includegraphics[width=\columnwidth]{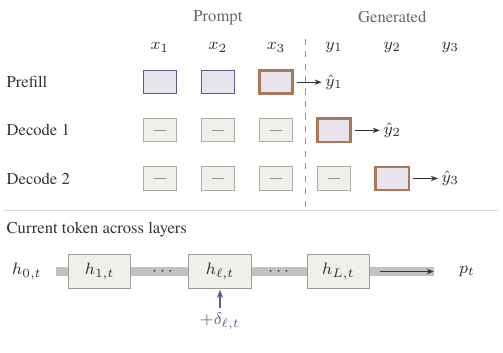}
\caption{KV reuse and edits to the current token across depth. Purple cells are newly computed keys and values; gray cells with a dash are reused. Brown outlines mark the current position. Arrows predict $\hat y_i$; sampled $y_i$ becomes an input on the next step. Below, the current token is expanded across layers, with $\delta_{\ell,t}$ marking a possible edit at one selected layer; edits may target one or several layers.}\label{fig:timing}
\end{figure}

In a causal transformer, the last prompt position predicts the first output token. After that token is sampled, it becomes the new input position for predicting the second. With KV caching, the model normally reuses past keys and values and computes the new position at each layer. An edit to the last prompt state is therefore different from edits to each new state or to every position in the prompt. Its persistence depends on whether the module edited can affect a key or value that later steps will read. Figure~\ref{fig:timing} makes the distinction explicit. The fact that a last position can attend to the entire available prefix establishes access, not faithful compression of every fact in it.

The phase in which a direction is estimated can differ from the phase in which it is applied. Source-preference contrasts at the final prompt position address the initial answer decision~\cite{lou2026ksd}. Contrasts estimated from selected positions can be deployed over a wider set of tokens~\cite{qiu2024sea}. Measuring shared directions in cached decoding and ablating them against prefill or random controls instead tests which directions support decisions while an answer unfolds~\cite{decodeshare2026}. In vision-language models, injecting a prompt-end embedding at later steps carries one summary forward. Modifying prompt-time keys and values carries an edited memory forward for different future queries~\cite{fazli2026cei,zhang2026pti}. Both can persist across tokens, yet only the latter changes the stored evidence read by attention.

A layer difference is meaningful only relative to a reference trajectory and an output of interest. Analyses of displacement across depth and studies of source preference show that a middle-layer signal can be predictive without deciding the final token~\cite{damirchi2026trajectory,residual2024conflict}. Damping a head update orthogonal to the current residual treats that residual as a local anchor~\cite{zhao2026dco}. Following a semantic setpoint with a local feedback model instead makes deviations from the specified target actionable~\cite{skifstad2026alqr}. Confidence-gradient updates and divergence-triggered compensation use the model's current state to determine how much to move~\cite{liu2026usteer,yu2026dac}. Each definition of error has a failure mode. Useful new evidence may arrive orthogonally to an old belief, a setpoint may encode the wrong source, and a confident model may be confidently wrong.

\section{Claim Verification and Selective Answering}\label{sec:verification}

\begin{figure*}[!b]
\centering
\includegraphics[width=\textwidth]{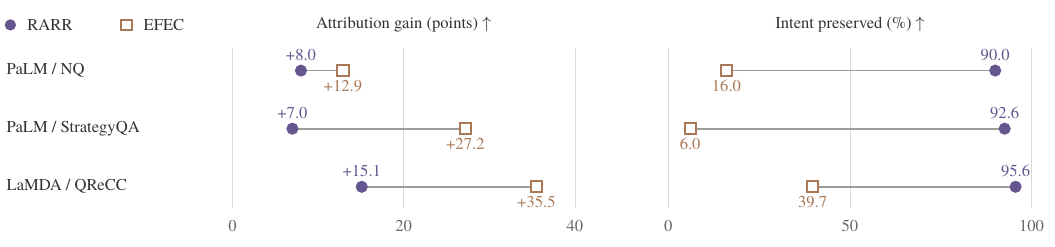}
\caption{Revision of the same drafts using shared retrieved evidence (150 test passages per condition). Rows identify the draft model and task. Left: change in mean sentence-level human attribution. Right: the share of drafts whose original intent is fully preserved. Filled circles denote RARR and open squares EFEC; lines pair the two editors within each condition. Values are from Table 1 of Gao et al.~\cite{gao2023rarr}.}\label{fig:revision}
\end{figure*}

A draft makes explicit the claims a system is about to endorse. At that point it can change the content of a particular claim, seek evidence before keeping it, or withhold the answer. These actions have different costs: a successful correction preserves information, while a well-placed abstention avoids an unsupported commitment. Both can raise measured accuracy under different denominators.

\subsection{Independent evidence and local revision}

Revision can change an answer without access to model weights or token probabilities. Its value depends on what the second pass knows that the first did not. Self-Refine uses self-generated feedback that may expose a contradiction or missing step, while Chain-of-Verification (CoVe) answers verification questions separately to reduce dependence on the wording of the initial draft~\cite{madaan2023selfrefine,dhuliawala2024cove}. Both still draw on the same model memory. Tests of intrinsic self-correction on reasoning tasks found that another attempt without external feedback can preserve or introduce errors~\cite{huang2024selfcorrect}. This is a reason to measure corrections and regressions separately, and to distinguish repeated attempts from the addition of a genuinely new source.

External checks can target an uncertain reasoning step or a claim in a completed response. Searching for evidence about the disputed step can change subsequent reasoning~\cite{zhao2023verifyedit}. RARR locates unsupported claims and makes evidence-grounded revisions while trying to preserve what the author meant~\cite{gao2023rarr}. A learned detector/editor can target a finer error taxonomy, at the cost of training the editing component~\cite{mishra2024fava}. By contrast, selecting short supporting spans before writing each sentence makes some later verification cheaper, but a missed span can prevent a valid fact from appearing in the first place~\cite{slobodkin2024attribute}. Thus verification timing trades freedom to synthesize against the cost of tracing an already-written claim back to its evidence.

Figure~\ref{fig:revision} compares two editors that received the same retrieved evidence. On Natural Questions, the supervised editor EFEC raises the attribution score by 12.9 points and preserves the original intent in 16.0\% of cases. RARR gains 8.0 points and preserves intent in 90.0\%. The same ordering holds when the inputs are reasoning chains or conversational answers~\cite{gao2023rarr}. Higher attribution can accompany substantial changes to what a user asked the draft to convey. An evaluation of revision should therefore report both newly supported claims and whether the supported substance of the original answer survived~\cite{gao2023rarr,gao2023alce}.

\subsection{Uncertainty, answerability, and abstention}

Repeated samples provide a black-box signal. Comparing their claims measures agreement, and grouping semantically equivalent answers discounts differences that are merely paraphrases~\cite{manakul2023selfcheck,kuhn2023semantic,farquhar2024entropy}. With hidden states available, the spectrum of sample representations provides another measure of variation~\cite{chen2024inside}. A model's confidence in a proposed answer and its confidence that it can answer a question are also distinct judgments~\cite{kadavath2022know}. All can help prioritize verification. A widely memorized misconception, however, can be repeated consistently, stated confidently, and expressed by a compact cluster. Agreement across samples from one model gives less independent evidence than agreement with a well-sourced external record.

Acting on a risk score requires a policy. A trained attention-ratio classifier can select among sampled continuations~\cite{chuang2024lookback}. Black-box prompts can explicitly allow abstention or ask the model to compare verbal confidence with a threshold~\cite{madhusudhan2025abstain}. A calibrated similarity threshold offers a way to control a specified risk under the method's sampling assumptions~\cite{abbasi2024conformal}. These actions consume different resources and discard different material. When the information missing is one date, a second search or a qualified partial answer may retain useful content. When the question's premise is unsupported, abstaining with the reason can be the correct outcome. Controlled perturbations of grounded questions make both mistaken refusal and failure to refuse measurable on matched tasks~\cite{muhamed2026refusalbench}.

Selective answering is valuable even when the proposition has a definite truth value. A document may omit the requested date while giving the location. Withholding the date and returning the supported location conveys more than refusing the whole question. Evaluation at claim level can score the two decisions separately. A whole-answer refusal rate cannot reveal whether this partial use of evidence succeeded~\cite{slobodkin2024attribute,muhamed2026refusalbench}.

Conformal factuality makes the amount retained an explicit decision~\cite{mohri2024conformal}. Its implementation splits a draft into subclaims, scores their uncertainty, and backs off by removing less reliable claims. Calibration chooses a threshold for the resulting level of specificity. The guarantee is marginal over exchangeable calibration and test examples, relative to the reference entailment judgment; it is not certainty for each answer. This T1 policy differs from whole-answer abstention and from CDA's token-level refusal branch. Evaluation must measure retained information as well as correctness.

\section{Benchmarks and Evaluation}\label{sec:evaluation}

\subsection{Units of judgment and sources of ground truth}

Truthfulness, grounding, and detection are evaluated at different units. TruthfulQA tests questions designed around common misconceptions and distinguishes open generation from multiple-choice evaluation~\cite{lin2022truthfulqa}. HaluEval asks whether a supplied item contains an error, whereas RAGTruth annotates unsupported text within responses generated from supplied evidence~\cite{li2023halueval,niu2024ragtruth}. FActScore checks atomic facts in long answers. Search-assisted long-form evaluation can also account for the number of supported facts produced~\cite{min2023factscore,wei2024longfact}. A model that drops difficult statements can improve the supported fraction while conveying less information. In citation-based answers, a source can cover only part of a long sentence or be redundant with another citation~\cite{gao2023alce}. A single number cannot identify which behavior changed.

Table~\ref{tab:benchmarks} groups benchmarks by the object a score concerns. A detector tested on supplied responses directly establishes its ability to identify errors under that sampling scheme. A generator must also decide how many claims to make and when to stop. To transfer a detector result to a generation claim requires evaluating the text produced after the intervention, including correct statements removed in the process~\cite{li2023halueval,niu2024ragtruth,min2023factscore}.

Visual grounding needs two further units. POPE asks whether a named object is present; random, popular, and adversarial negative sampling create different tests of absent-object errors~\cite{li2023pope}. Report precision, recall, F1, and the fraction of ``yes'' responses with the sampling setting. CHAIR instead scores freely generated captions: $\mathrm{CHAIR}_i$ is the fraction of mentioned objects that are hallucinated, and $\mathrm{CHAIR}_s$ the fraction of captions containing a hallucinated object~\cite{rohrbach2018chair}. The original implementation uses MSCOCO object categories, synonyms, and image/caption annotations. Neither metric covers every attribute or relation. Shorter descriptions may reduce hallucinations while omitting useful objects, so caption length and supported content should accompany the score.

\begin{table*}[!t]
\centering
\caption{What a benchmark judges and what its score can support}\label{tab:benchmarks}
\footnotesize\renewcommand{\arraystretch}{1.16}
\begin{tabularx}{\textwidth}{@{}P{.204\textwidth}YY@{}}
\toprule
\textbf{Benchmark} & \textbf{Judgment and reference} & \textbf{Interpretation} \\
\midrule
TruthfulQA / FreshQA~\cite{lin2022truthfulqa,vu2024freshllms} & Answer to a misconception-prone question / answer at a specified time or to a false premise. & Truth and informativeness can move independently. FreshQA also depends on the chosen judging mode. \\
HaluEval / RAGTruth~\cite{li2023halueval,niu2024ragtruth} & Judgment on a supplied item / annotation of unsupported spans in generated responses. & Useful for detection and localization. Improving end-to-end generation requires a separate test. \\
FActScore / LongFact~\cite{min2023factscore,wei2024longfact} & Atomic claims checked with a knowledge source or search-assisted evaluation. & Supported fraction and number of supported claims answer different questions about a long answer. \\
ALCE~\cite{gao2023alce} & Statement support and citation contribution, alongside answer quality. & A citation may be relevant yet support only part of a statement. \\
Abstain-QA / RefusalBench~\cite{madhusudhan2025abstain,muhamed2026refusalbench} & Answerable and unanswerable questions under the respective task definitions. & Track both unwarranted answers and avoidable refusals, together with coverage. \\
POPE~\cite{li2023pope} & Yes/no judgments on object presence; specified negative-object sampling. & Sampling difficulty and answer bias affect the score; it is not a free-caption test. \\
CHAIR~\cite{rohrbach2018chair} & Hallucinated object mentions and captions containing them, relative to object/caption annotations. & Interpret alongside supported content and length; object scores do not measure every visual relation. \\
\bottomrule
\end{tabularx}
\end{table*}

\subsection{Reading a reported improvement}

For selective answering, let $A$ denote an answer and $E$ an error. Equation~(\ref{eq:risk}) separates coverage, error among answered questions, and error per submitted question. Conditional error is defined only at nonzero coverage. A method with lower joint risk may simply answer fewer questions. Threshold-dependent guarantees must therefore name the risk controlled and the assumed relationship between calibration and deployment data~\cite{abbasi2024conformal,mohri2024conformal}. Paired answerable and unanswerable sets measure a complementary error: rejecting a question that the supplied documents actually resolve~\cite{madhusudhan2025abstain,muhamed2026refusalbench}. For temporal questions, scoring only the main answer versus scoring every claim in the answer yields another difference in denominator~\cite{vu2024freshllms}.

\begin{equation}
\begin{aligned} C&=\Pr(A),\quad R_{\mathrm{sel}}=\Pr(E\mid A),\\ R_{\mathrm{joint}}&=\Pr(E\cap A)=C\,R_{\mathrm{sel}}.\end{aligned}\label{eq:risk}
\end{equation}

\begin{figure}[!t]
\centering
\includegraphics[width=\columnwidth]{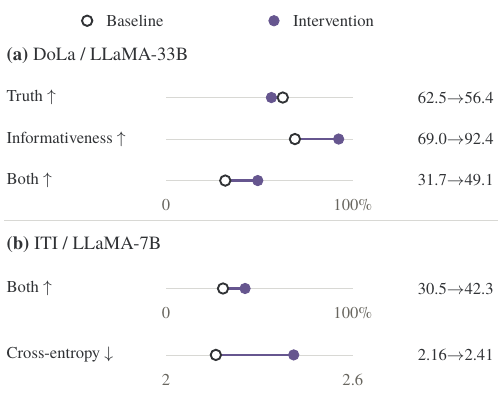}
\caption{Within-study comparisons. (a) DoLa: open-ended TruthfulQA with LLaMA-33B (Table 1). (b) ITI: LLaMA-7B in two-fold evaluation (Table 3). Arrows indicate the preferred metric direction. Truth, informativeness, and joint scores are percentages; OpenWebText cross-entropy uses its own scale. Each pair shares a protocol within its source study~\cite{chuang2024dola,li2023iti}.}\label{fig:evidence}
\end{figure}

Figure~\ref{fig:evidence} illustrates why even results on one benchmark need their component scores. In the reported LLaMA-33B open-ended TruthfulQA comparison, layer contrast raises the fraction judged both true and informative from 31.7 to 49.1, while truthfulness alone falls from 62.5 to 56.4 and informativeness rises from 69.0 to 92.4~\cite{chuang2024dola}. In a separate LLaMA-7B experiment, a head intervention raises the joint score from 30.5 to 42.3 but increases OpenWebText cross-entropy from 2.16 to 2.41~\cite{li2023iti}. Search can produce a different pattern. On the original FreshQA evaluation, adding FreshPrompt to GPT-4 raises strict accuracy by 47.0 percentage points and relaxed accuracy by 31.4~\cite{vu2024freshllms}. These comparisons identify different changes within each paper's setup. Their magnitudes cannot rank the three families.

\subsection{Capability preservation and comparison budgets}

Capability preservation requires its own outcome. Removing a shared decoding subspace can impair several tasks, but the effect of ablating an entire space need not predict the effect of moving a small distance in it~\cite{decodeshare2026,sharma2026lowrank}. A large geometric overlap between two behavioral spaces likewise does not fix the size or symmetry of their interaction~\cite{sharma2026lowrank}. Measuring a target behavior at several operating points under a capability-loss budget exposes trade-offs hidden by a single strength setting~\cite{raj2026frontier}. Tests of steering specificity and truthfulness versus safety further show why unrelated benchmarks, nearby control behaviors, and shifted inputs require separate checks~\cite{specificity2026,safetytradeoff2026}. Faithfully acknowledging a cue that influenced a reasoning trace is another meaningful behavior, with different ground truth from factual or source-faithful generation~\cite{nguyen2026cotfaith}.

Inference budgets also change what the comparison can establish. A second decoding branch spends work on every treated step. A verification system may instead spend most of its budget on a few expensive searches. A steering vector may be cheap to apply after an expensive layer and strength sweep. An interpretable report distinguishes calibration from deployment, discloses extra calls and generated tokens, and compares approaches at matched answer coverage or latency where the task permits.

\section{Application Settings}\label{sec:applications}

Three settings illustrate how the standard of evidence changes within a single intervention family. In current-fact QA the world at the date of the question supplies the reference. In document synthesis the stipulated record defines what may be asserted. In visual grounding the visible scene constrains what can be said about an object. The same tendency to follow context can be helpful in one setting and misleading in another.

\subsection{Time-sensitive and multi-hop question answering}

Suppose a user asks who currently heads an institution. A document naming a former head can be topically perfect and temporally wrong. Retrieval should expose a dated authoritative source. Context-sensitive decoding can help a frozen model express an unfamiliar new name once that source is available. If dated accounts disagree, grouping evidence by provenance before generation makes the conflict inspectable~\cite{vu2024freshllms,shi2024cad,wang2025astute}. Merely increasing reliance on the first result has the opposite effect when the older appointment ranks first.

\begin{figure}[!t]
\centering
\includegraphics[width=\columnwidth]{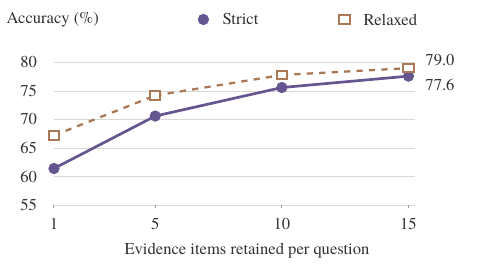}
\caption{GPT-4 with FreshPrompt on the original FreshQA test set (500 questions, five demonstrations). Points mark the four counts of evidence items retained in the prompt. Strict evaluation checks all assertions. Relaxed evaluation credits a correct main answer when other details do not contradict it. Source: Tables 1 and 6 of Vu et al.~\cite{vu2024freshllms}.}\label{fig:budget}
\end{figure}

In multi-hop QA the first retrieval may be factually sound yet answer the wrong subquestion. Exposing the bridge entity before issuing the next query allows each hop to be checked~\cite{press2023selfask,trivedi2023ircot}. FreshQA highlights a different failure after the correct entity has already been found. Its relaxed judgment can credit a correct primary answer despite an inaccurate addition, whereas the strict judgment also scrutinizes that addition~\cite{vu2024freshllms}. An answer containing the right name and a fabricated appointment date illustrates the difference. Recording retrieved evidence, the primary answer, and its accompanying claims distinguishes these two sources of error.

The number of evidence items kept in the prompt also affects accuracy. In the original FreshQA experiment, holding the model and demonstrations fixed while increasing the number of retained items from one to five raises strict accuracy by 9.2 points. Going from ten to fifteen adds 2.0 points (Fig.~\ref{fig:budget}). The four measurements show how the observed gain depends on the amount of evidence already in the prompt. They do not identify an optimal number of retained items for other dates or collections~\cite{vu2024freshllms}.

\subsection{Document synthesis and attributed answers}

In a document summary, adding a plausible cause to a reported event can introduce an unsupported claim even when every named entity is correct. Local attribution addresses this before writing by selecting source spans and grouping those that jointly license a sentence~\cite{slobodkin2024attribute}. Post-generation revision instead searches for support for claims already made and edits unsupported material~\cite{gao2023rarr}. The first approach makes unsupported elaboration easier to locate at drafting time. The second can repair legacy drafts or outputs from a closed model. Either can lose useful content: span selection can omit a qualifier, and a reviser can replace a supported clause while fixing its neighbor.

A sentence with two clauses and two sources is a useful test. The cited set should jointly support both clauses. Each citation should also contribute support to what was written, and the answer should retain the facts requested in the prompt. These are separable judgments in citation-based evaluation~\cite{gao2023alce}. Automated entailment adds measurement error: an evaluator may accept two topically similar sentences despite different dates. Mixing human checks of such borderline cases with claim coverage and citation-level evidence is more informative than counting references per answer~\cite{gao2023alce,min2023factscore}.

\subsection{Visual grounding and ambiguous observations}

Images make it possible to perturb one evidence channel while keeping the linguistic question constant. Contrasting intact and degraded images estimates dependence on visual input~\cite{leng2024vcd}. Intermediate visual readouts test whether evidence visible earlier is lost before prediction~\cite{wang2025deco}. Contrasting two activation-steered branches combines an internal edit with output rescoring~\cite{su2025asd}. Generating a caption first creates a textual proxy for the image whose reliability depends on the question~\cite{li2026geass}. These are four distinct hypotheses about the source of the same wrong object word. Attention penalties with rollback address repetitive overreliance on particular prior tokens, while instruction-preserving decoding tests whether an intervention also removes the model's ability to decline a question~\cite{huang2024opera,zhang2026instruction}. A text analogue requires an explicit passage perturbation and its own grounding test.

Visual studies make the object of an edit particularly explicit. One can estimate visual versus language-prior contributions to heads with trained potential functions, use local sensitivity or gradient-estimated route effects to select heads, or use an analytic energy gradient to redistribute attention away from sink tokens and toward the image~\cite{chang2026ahead,ji2026causallens,cheng2026route,luo2026lad}. Other methods estimate a space from representations before and after visual attention drops, or partition a state into visual, prior, and residual components~\cite{yin2026rivs,lin2026hulluedit}. Cluster-specific directions select among a bank of edits, while per-layer ridge regressions fitted to paired source-preference scores yield a distinct kind of steering vector~\cite{yin2026dynamic,wu2026cas}. The latter fits a predictive model even though only its coefficients are used to steer. These approaches all change visual grounding, but deleting a suspected prior, restoring lost evidence, and reallocating attention predict different failure modes. When the image is ambiguous, successful suppression of a language prior can also remove a useful inference.

For a caption, naming an object that disappears when the image channel is perturbed is evidence of model sensitivity to the image. It does not settle whether that object was correctly identified. A cropped region might contain an unrecognizable object, and an annotator may distinguish “absent” from “not visible.” Evaluating object mentions alongside omissions and the ability to decline an unanswerable question separates better grounding from indiscriminate suppression of object words~\cite{leng2024vcd,zhang2026instruction,sahoo2024survey}. In text, an analogous test would distinguish a document that contradicts a claim from one that simply does not mention it.

This is why the visual methods above should be tested on both object-presence questions and generated descriptions. POPE exposes false affirmations under controlled object sampling, whereas CHAIR measures invented object mentions in the caption actually produced~\cite{li2023pope,rohrbach2018chair}. Agreement between the two is more informative when correct object coverage and the ability to decline remain intact.

\section{Open Research Questions}\label{sec:challenges}

\subsection{Source reliability under knowledge conflict}

The first unresolved question is how a system should choose its authority under conflict. Experiments that reward following any supplied passage teach a different behavior from experiments that require rejecting stale or misleading evidence~\cite{longpre2021knowledge,xie2024conflict,wang2025astute}. A more informative test pairs the same question with verified updates, plausible false updates, insufficient evidence, and an explicit instruction to answer within a counterfactual document. Annotating which source should prevail lets an evaluator distinguish a correct source switch from a model becoming easier to persuade. The task must also specify when the evidence warrants a qualified answer.

An informative design would additionally separate the authority of the source from how persuasively its claim is written. Dates, independent provenance, and the task instruction can be varied without changing the candidate answers. The dependent variable is then whether the model switches on the cases where switching is warranted, with false switches on the matched controls reported alongside. This would test whether source-sensitive steering can improve decisions beyond encouraging compliance with any retrieved account~\cite{lou2026ksd,xie2024conflict,wang2025astute}.

\subsection{Intervening on trajectories that can recover}

A grounding threshold can trigger a minimum-size correction, while a trained flow can learn a different correction for each query~\cite{jana2026bracs,wang2025truthflow}. Dynamic behavior is compatible with either training regime. To test a proposed trajectory intervention, a paired design could first establish which unedited runs recover after an intermediate deviation. It could then apply the same edit to matched recovering and persistently wrong runs, observing downstream answers, later states, and collateral errors. An intervention that keeps all layers near a reference while turning recoveries into mistakes has optimized the chosen trajectory metric at the expense of the answer. This is a proposed discriminating test, distinct from the experiments reported by the cited work.

\begin{figure}[!t]
\centering
\includegraphics[width=\columnwidth]{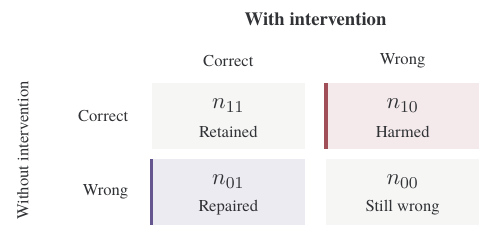}
\caption{Paired outcomes in a proposed trajectory test. First select prompts with a prespecified mid-layer deviation on the unedited run. For each prompt, score the final answer with and without the edit. Rows and columns are those two final judgments ($1$ correct, $0$ wrong). Cell symbols denote counts to be collected.}\label{fig:transitions}
\end{figure}

Figure~\ref{fig:transitions} makes the unit of comparison explicit. Hold the prompt, evidence, decoding rule, and answer key fixed. Predefine the deviation criterion and reference on held-out calibration data, then select cases using only the unedited trajectory. On the resulting $N$ paired questions, the accuracy change is $(n_{01}-n_{10})/N$. Among initially wrong answers the repair rate is $n_{01}/(n_{00}+n_{01})$. Among initially correct answers the harm rate is $n_{10}/(n_{10}+n_{11})$. Each rate requires a nonempty denominator. Equal net gains can hide very different repair and harm rates.

Position and depth should also be varied separately. Prompt-final edits change the state used for the first prediction. Decode-current edits repeatedly treat the position that predicts the next token. Editing all prompt positions can alter stored evidence before later queries are known. A matched comparison would keep the calibration set, layer window, and displacement budget fixed, record the cache policy, and score both the first disputed entity and factual consistency later in the answer~\cite{qiu2024sea,lou2026ksd,decodeshare2026,zhang2026pti}.

\subsection{Degrees of freedom and collateral errors}

There is little reason to expect one extra subspace dimension to have the same effect across tasks. It might separate the target from a nuisance feature, allow a new state-dependent displacement, or expose a direction shared by ordinary decisions~\cite{xenos2026ccs,decodeshare2026,sharma2026lowrank}. These possibilities can be distinguished by comparing a projected fixed vector with variable coefficients in the same basis and a version constrained by a specified readout~\cite{luo2026skop}. Matching the basis, displacement norm, and calibration split makes the operator the main difference. Reporting which previously correct examples turn wrong would test whether a higher average target score was purchased by sacrificing a particular kind of knowledge.

\section{Conclusion}\label{sec:conclusion}

Training-free hallucination intervention is best understood by following evidence to the decision it can still change. Search supplies facts absent from the prompt. Decoding and representation control alter how existing evidence contributes to a prediction. Verification and abstention act when a claim can be examined or when support is inadequate. Across these choices, three observations change the interpretation of a gain: whether the preferred source is authoritative for the task, whether an intervention preserves information that later decisions need, and whether correctness improved at comparable information content and answer coverage. For activation steering in particular, the decisive test is whether a specified change to a layer, token, and subspace improves source selection while preserving previously supported answers.

\appendices
\section{Review Scope and Evidence Selection}\label{sec:method}

This structured scoping review covers public literature available by November 20, 2026, with a targeted check on November 23. Searches combined hallucination, factuality, faithfulness, and knowledge conflict with retrieval, contrastive decoding, activation steering, subspace, verification, and abstention. Sources included ACL Anthology, NeurIPS proceedings, OpenReview, PMLR, CVF, IJCAI, publisher records, and arXiv, supplemented by citation tracing and exact-title checks. The follow-up combined ``training-free'' with verification, correction, and abstention and checked the original POPE and CHAIR records.

We included concrete inference-time corrections and analyses or evaluations needed to interpret them. General steering is labeled enabling evidence when hallucination is not tested. Trained interventions remain comparators outside the T0/T1 core. Preprints and proceedings versions share one record; unverifiable sources and unsupported publication claims are excluded. The corpus contains 116 works, including 44 core methods (32 T0, 12 T1) and 21 primarily vision--language studies. Figures~\ref{fig:corpus} and~\ref{fig:preparation} describe this selected corpus, not field-wide publication prevalence. Nineteen works are cited as arXiv preprints; one reports NeurIPS 2026 Spotlight acceptance in its November 19 revision, while the other 18 have no confirmed acceptance or publication in the checked cutoff records.

For each record, we checked the primary bibliographic source and passages relevant to its role; reading depth varies by paper. Numerical comparisons use original reports. The source-to-section index and primary URLs accompany the paper. Selection emphasizes publicly documented mechanisms and retains more white-box than answer-policy studies. The search log does not capture every initial hit, so neither screening counts nor retrieval recall are reconstructed. Differences in tasks, model versions, evidence, and outcome definitions preclude a pooled effect size.

\bibliographystyle{IEEEtran-survey}
\bibliography{references}
\end{document}